# Hybrid Inception Architecture with Residual Connection: Fine-tuned Inception-ResNet Deep Learning Model for Lung Inflammation Diagnosis from Chest Radiographs

Mehdi Neshat[a,b,c,*], Muktar Ahmed[b,c], Hossein Askari[d], Menasha Thilakaratne[e], and Seyedali Mirjalili[a]

[a]*Center for Artificial Intelligence Research and Optimisation, Torrens University Australia, Brisbane, QLD 4006, Australia.*
*mehdi.neshat@torrens.edu.au , ali.mirjalili@torrens.edu.au*
[b]*South Australian Health and Medical Research Institute (SAHMRI), University of South Australia, Adelaide, SA, 5000, Australia,*
*muktar.ahmed@unisa.edu.au*
[c]*Australian Centre for Precision Health, the University of South Australia Cancer Research Institute, Adelaide, SA, 5000, Australia*
[d]*School of Electrical Engineering and Computer Science, University of Queensland, St Lucia, Brisbane, QLD, 4072, Australia,*
*h.askari@uq.edu.au*
[e]*School of Computer Science, The University of Adelaide, Adelaide, 5005, Australia,*
*menasha.thilakaratne@adelaide.edu.au*

**Abstract**

Diagnosing lung inflammation, particularly pneumonia, is of paramount importance for effectively treating and managing the disease. Pneumonia is a common respiratory infection caused by bacteria, viruses, or fungi and can indiscriminately affect people of all ages. As highlighted by the World Health Organization (WHO), this prevalent disease tragically accounts for a substantial 15% of global mortality in children under five years of age. This article presents a comparative study of the Inception-ResNet deep learning model's performance in diagnosing pneumonia from chest radiographs. The study leverages Mendeley's chest X-ray images dataset, which contains 5856 2D images, including both Viral and Bacterial Pneumonia X-ray images. The Inception-ResNet model is compared with seven other state-of-the-art convolutional neural networks (CNNs), and the experimental results demonstrate the Inception-ResNet model's superiority in extracting essential features and saving computation runtime. Furthermore, we examine the impact of transfer learning with fine-tuning in improving the performance of deep convolutional models. This study provides valuable insights into using deep learning models for pneumonia diagnosis and highlights the potential of the Inception-ResNet model in this field. In classification accuracy, Inception-ResNet-V2 showed superior performance compared to other models, including ResNet152V2, MobileNet-V3 (Large and Small), EfficientNetV2 (Large and Small), InceptionV3, and NASNet-Mobile, with substantial margins. It outperformed them by 2.6%, 6.5%, 7.1%, 13%, 16.1%, 3.9%, and 1.6%, respectively, demonstrating its significant advantage in accurate classification.

1. **Introduction**

Accurate lung inflammation (pneumonia) diagnosis is paramount for successful treatment and management. Pneumonia is a prevalent respiratory infection that can be caused by bacteria, viruses, or fungi and can impact individuals of all ages. The World Health Organization reports that pneumonia accounts for 15% of all deaths of children under five years old worldwide and affects millions of adults annually [1]. The mortality rate attributed to pneumonia is influenced by factors such as age, overall health, and the type of microorganism causing the infection. Early and precise diagnosis of pneumonia is critical to decrease the morbidity and mortality rates linked with this disease.

Accurately diagnosing pneumonia is critical because a prompt and correct diagnosis can lead to timely treatment, improving patient outcomes and reducing the risk of complications [2]. Secondly, accurate diagnosis can assist in preventing the disproportionate service of antibiotics, which can contribute to the development of antibiotic resistance. Furthermore, it can speed up identifying pneumonia outbreaks and support public health efforts to control the spread of the disease. Nevertheless, there exist some significant challenges in accurately diagnosing pneumonia. One of the main challenges is the lack of specificity of the symptoms associated with pneumonia, which can overlap with other respiratory illnesses [3]. Besides, chest X-rays and different imaging approaches can be limited in resource-limited settings where trained personnel and equipment may not be available. Another challenge is distinguishing bacterial pneumonia from viral pneumonia, as they can have various treatment approaches. Last but not least, diagnosing pneumonia with high precision can largely depend on the experience and skill of the healthcare providers, which can vary across different settings.

Hashmi's study [4] used a cutting-edge deep learning technique for pneumonia detection that incorporated pretrained convolutional architectures such as ResNet18, Xception, InceptionV3, DenseNet121, and MobileNetv3 and introduced a weighted classifier for transfer learning via fine-tuning. This approach was coupled with partial data augmentation methods, and the model's performance was evaluated on the Guangzhou Women and Children's Medical Center pneumonia dataset, resulting in an impressive testing accuracy of 98.4% and an outstanding AUC score of 99.8% compared with other CNN models. In another study, Aledhari [5] proposed a CNN model and corresponded with pre-trained models such as ResNet50 and InceptionV3 to diagnose pneumonia. The proposed CNN model outperformed the pre-trained models in terms of accuracy, with an average accuracy of 64% and specificity of 69%. The model achieved 51% accuracy when utilising the VGG16 pre-trained model. While the proposed technique surpassed the current state-of-the-art algorithm VGG16, it still suffers from poor accuracy for medical image identification. In one recent study [6], an effective algorithm was introduced for localising lung opacities regions using the singleshot detector RetinaNet with Se-ResNext101 encoders that were pre-trained on the ImageNet dataset. The model's accuracy was enhanced by adding global classification output, applying heavy data augmentations, and using an ensemble of 4 folds and various checkpoints to generalise the model. The proposed approach achieved one of the best results regarding accuracy classification. Emtiaz et al. [7] proposed a new convolutional neural network model, called CoroDet, to detect Covid-19 from Chest X-ray and CT Scan images. CoroDet also classified Pneumonia images into three classes in the second diagnostic, where viral and bacterial Pneumonia was classified in the third diagnostic. The accuracy achieved by the model in the second diagnostic was 94.2%, and in the third diagnostic, it was 91.4%, outperforming the current state-of-the-art methods. Sourab et al. proposed a CNN architecture with 22 layers in their study [8], which utilized three different machine learning techniques (Support Vector Machine, Random Forest Classifier, and K-Nearest Neighbor) to extract and classify the CNN model's learned features. The dataset used in the study was Mendeley Data v2, consisting of 5856 chest x-ray images in both Pneumonia and Normal classes. Specific data augmentation techniques were applied to increase the dataset's variability, and overfitting was prevented using a regulariser in the first dense layer. The implementation of RF, KNN, and SVM classifiers using CNN model-trained features resulted in accuracies of 99.52%, 96.55%, and 97.32%, respectively. The proposed CNN-RF hybrid method achieved an accuracy of 99.52% and an AUC score of 98.7%, comparable to other conventional methods.

Although numerous deep learning-based methods have been developed and applied to predict and classify pneumonia disease, recent improvements in CNNs (such as the Inception-Resnet model) with more effective performance and lower computational runtime uplift this motivation to analyse the performance of the advanced CNNs for diagnosing pneumonia with chest CT scan images. In this study, the primary contributions are as follows.

- We develop a deep comparative framework of Inception-ResNet as well as several modern CNN-based architectures with transfer learning and fine-tuning techniques in pneumonia detection tasks. We train, test, and compare Inception-ResNet and other models using chest CT scan images.

- Due to the criticality of this task and the need to identify lung cancer with minimal error, we aim to determine the most robust CNN models and measure the effectiveness of transfer learning in classification.

- Furthermore, the importance of hyper-parameters, learning rate and batch size on the performance of InceptionResNet evaluated and optimal settings are suggested based on this case study.

The remainder of this article is structured as follows. Section 2 describes the details of Inception-ResNet architecture, and the dataset used. In the following, the experimental and comparison results are fully discussed and overviewed in Section 3.1. Ultimately, Section 4 summarises the findings and supplies prospective plans.

2. **Methodology**

2.1. *Dataset*

The dataset utilised in this study collected and published by the University of California San Diego, entitled '*Labeled Optical Coherence Tomography (OCT) and Chest X-Ray Images for Classification*' at Mendeley Dataset V2 [9] in 2018, which comprises 5856 Chest X-ray 2D images of Normal and Pneumonia classes. The Pneumonia dataset includes both Viral and Bacterial Pneumonia X-ray images. The images were separated into training and testing sets in preparation for the proposed method. The dataset contained 1048 images for testing and 4808 images for training. The images in the dataset were of size 512*512 pixels but were reduced for further computation.

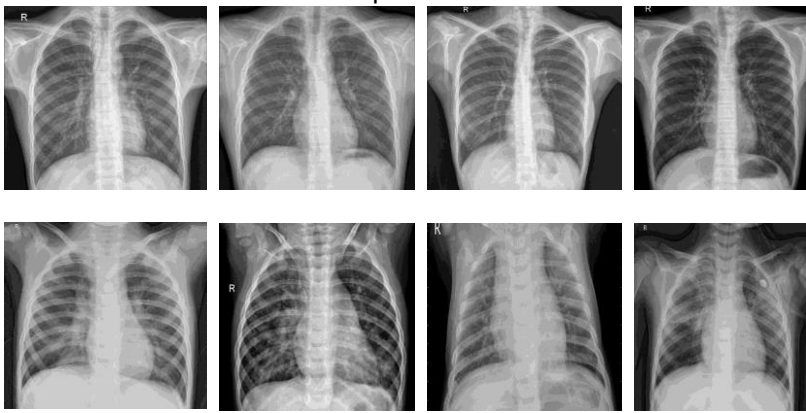

**Fig. 1: Some X-ray 2D images from the applied dataset. The first row shows four normal lungs, and the second row includes Pneumonia cases.**

Medical practitioners can diagnose pneumonia from X-ray images by identifying specific visual markers [10] in the image, such as infiltrates, consolidation, and air bronchograms. Infiltrates are unusual opacities in the lung tissue that may result from various substances, including fluid and pus. Consolidation, on the other hand, is the solidification of lung tissue due to the accumulation of exudate. Air bronchograms are air-filled bronchi visible against the opaque lung tissue. In addition to these indicators, the location and distribution of abnormalities in the lung are also considered. However, it is important to emphasize that the accuracy of diagnosing pneumonia from X-ray images may vary and should be confirmed by a qualified medical professional.

## 2.2. Inception-ResNet model

The InceptionResNetV2 architecture (described in [11]) combines the inception and residual connection models for improved performance. This hybrid approach enables the network to leverage the benefits of both models, including faster training times and avoidance of the vanishing gradient problem. The residual connections also allow the network to skip some layers during training. Additionally, InceptionResNetV2 employs multiple-size kernels in a single layer to yank patterns with diverse hierarchies, further enhancing the network's ability to catch features of varying complexity [12].

The Inception-ResNet model has multiple blocks containing convolutional layers, filter concatenation, ReLU activation functions, ResNet and inception structures. These blocks' architectural layout is depicted in Figure 2. The Inception-ResNetV2 model's distinctive design and effective use of filters have yielded impressive outcomes in medical image classification tasks. As a result, we employed the Inception-ResNetV2 pre-trained model as our foundation for this study.

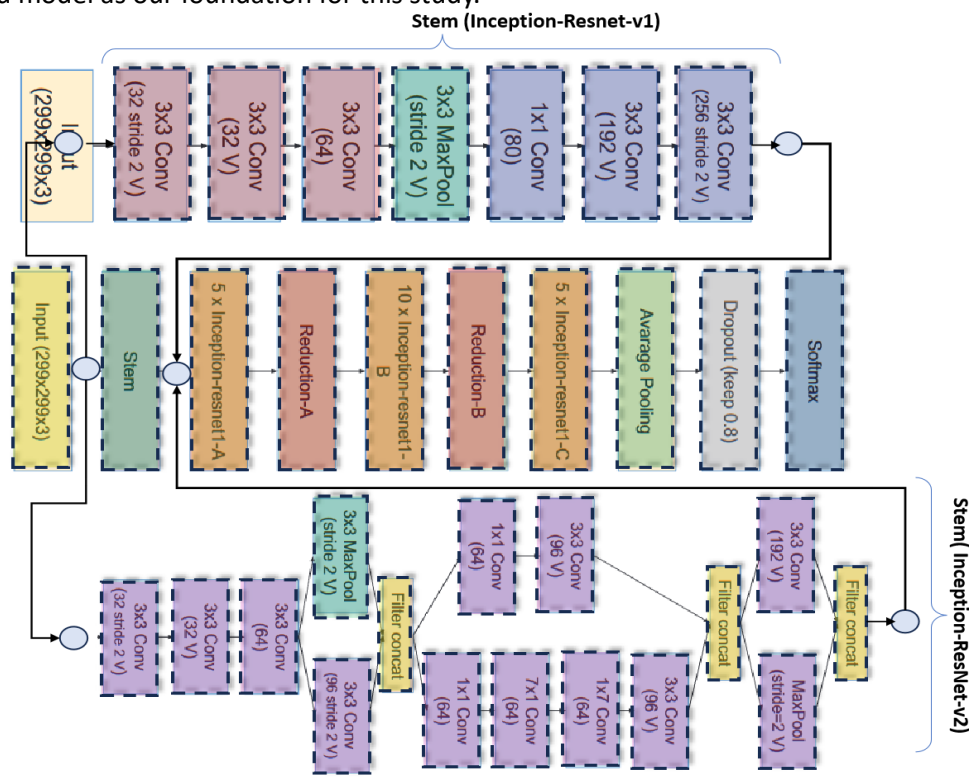

Fig. 2: Schematic of Inception-ResNet-v1 and v2 networks with the detailed stem sections.

The Inception-ResNet-v2 approach is commonly employed in CT-scan image classification [13, 12] owing to its ability to extract significant features from medical images. This deep learning architecture combines

Inception and ResNet, two well-known deep learning models. Inception models are recognized for their capability to extract features at multiple scales, whereas ResNet models effectively address the issue of vanishing gradients during training. The Inception-Resnet v2 approach is tailored to overcome the limitations of these models and attain high accuracy in image classification tasks [14]. Additionally, this approach has effectively handled large datasets and reduced computation time, making it a preferred choice for medical image analysis.

*2.3.* **Advanced Deep learning architectures**

ResNet152V2, proposed by He et al. [15], represents a family of highly deep architectures that exhibit impressive accuracy and favourable convergence properties. By examining the underlying propagation formulations within the residual building blocks, it is evident that the forward and backward signals can be efficiently propagated between blocks by utilizing identity mappings as skip connections and applying activation functions after addition. A comprehensive set of ablation experiments confirms the significance of these identity mappings. Based on this motivation, a novel residual unit, known as ResNet152V2, is introduced to facilitate training and enhance generalization capabilities. MobileNetV3, as proposed by Howard et al. [16], represents an enhanced iteration of MobileNetV2. It combines hardware-aware network architecture search (NAS) with the NetAdapt algorithm, further refined by introducing novel architectural advancements. MobileNetV3 serves as a prime example of the synergy between automated search algorithms and network design, leading to significant advancements in the field. Notably, MobileNetV3 demonstrates remarkable performance in tasks such as mobile classification, detection, and segmentation. In particular, MobileNetV3Large achieves a 3.2% increase in accuracy for ImageNet classification while simultaneously reducing latency by 20% compared to its predecessor, MobileNetV2. EfficientNetV2 [17] presents a fresh lineage of convolutional networks with notable improvements in training speed and parameter efficiency compared to earlier models. The development of this model family entailed a fusion of training-aware neural architecture search and scaling techniques to optimize training speed and parameter efficiency together. Experimental results demonstrate that EfficientNetV2 models exhibit significantly faster training times than previous models while also exhibiting a reduction in size by up to 6.8 times.

*2.4.* **Fine-tuning**

In this study, we used fine-tuning technique in order to improve the performance of large pre-trained convolutional models. Fine-tuning is an effective strategy used in large convolutional deep-learning models to acclimate a pretrained model to an unexplored dataset [18]. It concerns bringing a current model trained on an extensive dataset, such as ImageNet, and preparing it also on a smaller dataset to extract new features. This strategy can be utilised to leverage the understanding acquired from training on a large dataset and transfer it to the new problem. Indeed, it allows the model to learn fast and efficiently [19].

The fine-tuning process generally comprises pre-training and fine-tuning stages. Initially, an expansive convolutional neural network like MobileNetV3Large is trained on a large dataset (e.g., ImageNet) to learn a versatile set of features useful for various classification or regression tasks, particularly in vision. Subsequently, the pre-trained model is further trained on our dataset, adjusting its weights to suit the new data better while retaining the basic features learned during pre-training. However, fine-tuning involves the conscientious selection of hyper-parameters, such as the number of layers to be frozen (only a few top layers are set to be trainable to mitigate overfitting), learning rate, batch size, and the architecture of the pre-trained model. Striking a balance between conforming the model to the new task and preventing overfitting on the smaller dataset is crucial [20].

## 2.5. Transfer learning

Transfer learning (TL) [21] has emerged as a powerful technique, enabling the application of machine learning (ML) to real-world problems that traditional ML approaches struggle to handle. Various domains have successfully implemented TL [21]. TL primarily tackles three key challenges encountered in traditional ML: limited availability of labelled data, computational power constraints, and distribution mismatch. The essence of transfer learning lies in transferring the weights, or learned parameters, of a network from a source task (referred to as "task A") to a new target task (known as "task B"). Instead of initiating the learning process from scratch, we leverage the patterns and representations acquired from addressing a related task. By leveraging pre-existing knowledge, TL overcomes these hurdles and enhances the performance of ML models. TL has proven to be a valuable tool, addressing the limitations of traditional ML in various domains. Its ability to overcome the issues has paved the way for significant advancements in image processing, speech recognition, NLP, and beyond. Transfer learning techniques play a crucial role in enhancing the efficiency of deep learning models. They achieve this by reducing the time required for training, improving the utilization of available data, enhancing the model's ability to generalize, effectively handling the complexity of deep models, and mitigating the problem of overfitting. These advantages make transfer learning an invaluable approach for leveraging existing knowledge and achieving superior performance even when faced with limited resources.

## 3. Experimental results

### 3.1. Evaluation metrics

Evaluation metrics are a critical component of the ML workflow, enabling us to estimate the effectiveness of our proposed models and make data-driven determinations about how to enhance them compared with other existing ML techniques. In machine learning, various evaluation metrics are available that can be applied based on the task and objectives of the model. For example, metrics such as accuracy, precision, recall, F1 score, or area under the ROC curve (AUC) can be used to evaluate the model's performance for classification tasks. These metrics aid in comparing different models and selecting the best-performing model based on validation or test data. Additionally, they can be used to identify areas where the model needs improvement and guide efforts to adjust the model's hyper-parameters or features. Table 1 shows the details of evaluation metrics used in this study. True Positive (TP) represents the cases where the model predicted the positive class correctly. On the other hand, True Negative (TN) refers to the number of instances correctly classified as unfavourable by the model. The number of cases that are wrongly classified as positive by the model is represented by False Positive (FP). Finally, False Negative (FN) refers to the number of instances incorrectly classified as negative by the model and is also known as a Type II error.

## 3.2. Inception-ResNet performance analysis

In order to assess the performance of Inception-ResNet, we tested and compared it with several advanced CNN models. To take advantage of learned feature maps and avoid training a large model on a vast dataset from scratch, we utilised eight successful CNN models for transfer learning and fine-tuning purposes. These models include ResNet152V2, MobileNet-V3 (Large and Small), InceptionRes-NetV2, EfficientNetV2 (Large and small), InceptionV3 and NASNet-Mobile. Figure 3 demonstrates the effectiveness of the InceptionRes-NetV2 and other seven deep models in terms of accuracy (a), AUC (b) and loss validation (c).

In both accuracy and AUC, greater values show more desirable performance, and loss should be minimised. From Figure 3 (a), we can observe that Inception-ResNet-V2 outperformed ResNet152V2, MobileNet-V3 (Large and Small), EfficientNetV2 (Large and small), InceptionV3 and NASNet-Mobile at 2.6%, 6.5%, 7.1%, 13%, 16.1%, 3.9%, and 1.6%, recpectively, considerably in terms of classification accuracy. The second most significant observation is that NASNet-Mobile and ResNet152V2 performances are competitive with regard to accuracy and AUC. However, the average loss values of the ResNet152V2 method is high and cannot be a reliable model.

Monitoring the convergence rate of training and validating loss is an important aspect of training deep learning models for classification problems, as it can help us optimise the training process, improve the model's performance, and avoid overfitting. In this way, we can see a convergence speed comparison between all eight deep models for loss of training and validating in Figure 4 (a) and (b), respectively. Practically, NASNet-Mobile had the highest speed on training at the initial ten epochs; however, it could not keep this convergence and struggled with a local optimum. As illustrated in Figure 4 (a), this is clear that both Inception-ResNet-V2 and ResNet152V2 performed better than other models. Despite this, the loss validation convergence rate of ResNet152V2 is similar to InceptionV3 (See Figure 4(b)).

Table 1: The details of performance evaluation metrics for classifying Pneumonia cases. *TP*, *TN*, *FP*, and *FN* represent the four components of a confusion matrix.

| Abbreviation | full name | formula |
|---|---|---|
| TPR | True Positive Rate | $= \frac{TP}{TP+FN}$ |
| FPR | False Positive Rate | $= \frac{FP}{TN+FP}$ |
| PPV | Positive Predictive Value | $= \frac{TP}{TP+FP}$ |
| ACC | Accuracy | $= \frac{TP+TN}{TP+TN+FP+FN}$ |
| PRE | Precision | $= \frac{TP}{TP+FP}$ |
| REC | Recall | $= \frac{TP}{TP+FN}$ |
| SPF | Specificity | $= \frac{TN}{FP+TN}$ |
| AUC | Area Under the ROC Curve | $= \int_0^1 TPR(FPR^{-1}(t))\,dt$ |
| F1-score | Harmonic mean of PPV and recall | $= \frac{2*PPV*Recall}{PPV+Recall} = \frac{2*TP}{2*TP+FP+FN}$ |

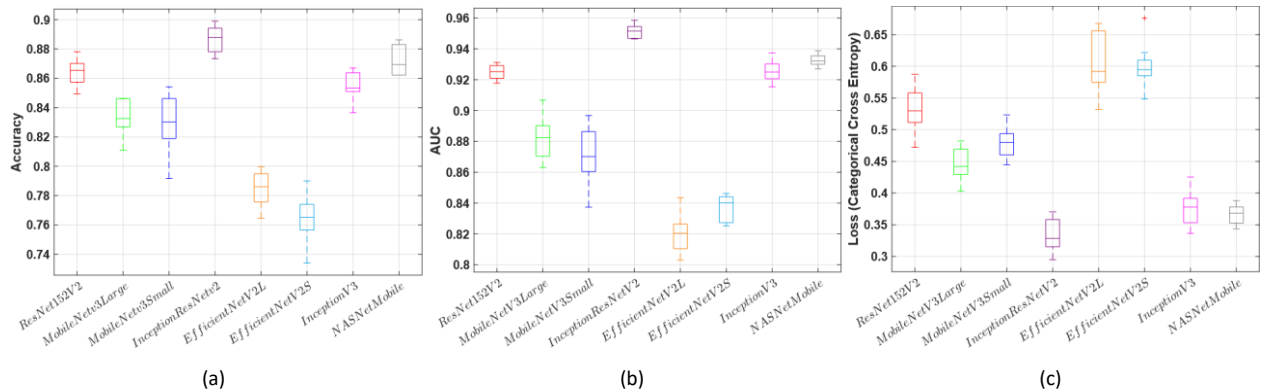

(a)                 (b)              (c)

Fig. 3: Statistical (a) Accuracy, (b) AUC, and (c) loss analysis for identifying normal and Pneumonia cases detection using Inception-ResNet-V2 compared with seven other modern Deep CNN models with transfer learning (weights='*imagenet*') and fine-tuning.

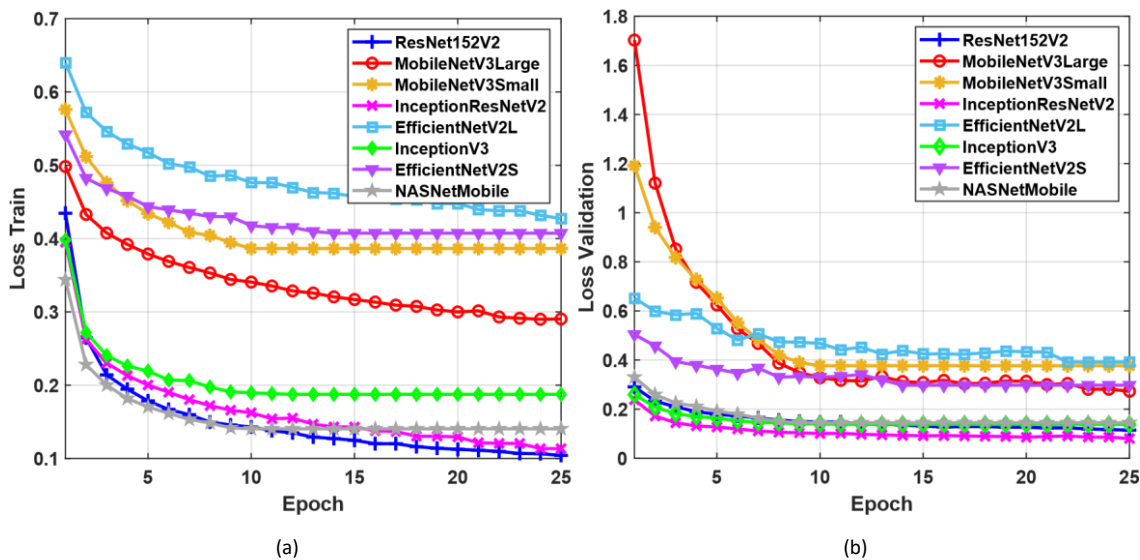

(a)                 (b)

Fig. 4: Convergence rate of (a) loss training, and (b) loss validation analysis for Inception-ResNet-V2 compared with seven other modern Deep CNN models with transfer learning (weights='*imagenet*') and fine-tuning.

Table 2 presents the five evaluation metrics' average scores for eight deep models and indicates that InceptionResNetV2 is the best-performing CNN model. It is worth noting that the top-performing model, Inception-ResNetV2, achieved the highest metrics for Pneumonia detection, including 88.7% accuracy, 0.864 Loss, 87.3% Precision, 0.95 AUC and 87.7% F1-score. The table's most significant observation is that designing a model with numerous classification layers like MobileNet-V3Large does not necessarily guarantee better performance than a classifier with a smaller training volume (Inception-ResNetV2) and may even exacerbate the overfitting issue. Furthermore, based on the recorded training runtimes for one epoch with a consistent number of trainable layers (TL=30), it is evident that the MobileNetV3Small model exhibits the fastest training speed, while the EfficientNetV2L model is the most computationally expensive. These findings are based on the following hardware information system: Processor: AMD Ryzen 9 6900HS with Radeon Graphics, operating at a frequency of 3301 MHz. It features eight cores and 16 logical processors, allowing for efficient parallel processing during training. Moreover, the system has 16.0 GB of physical memory (RAM). Graphics Processing Unit (GPU) has an NVIDIA GeForce RTX 3080 laptop

GPU. This powerful GPU enhances the training process by accelerating computations and optimizing performance.

Table 2: The statistical classification results of Inception-ResNetV2 compared with seven modern deep convolutional learning methods for ten independent runs with 5-fold cross-validation.

|  | ResNet152V2 | MobileNetV3Large | MobileNetV3Small | InceptionResNetV2 | EfficientNetV2L | EfficientNetV2S | InceptionV3 | NASNetMobile |
|---|---|---|---|---|---|---|---|---|
| Loss | 5.311E-01 | 4.466E-01 | 4.818E-01 | 3.332E-01 | 6.042E-01 | 5.995E-01 | 3.755E-01 | 3.663E-01 |
| Accuracy | 8.644E-01 | 8.325E-01 | 8.279E-01 | 8.869E-01 | 7.849E-01 | 7.630E-01 | 8.548E-01 | 8.724E-01 |
| Precision | 8.634E-01 | 8.200E-01 | 8.018E-01 | 8.727E-01 | 7.695E-01 | 7.600E-01 | 8.501E-01 | 8.720E-01 |
| AUC | 9.251E-01 | 8.817E-01 | 8.696E-01 | 9.515E-01 | 8.198E-01 | 8.377E-01 | 9.253E-01 | 9.327E-01 |
| F1-score | 8.644E-01 | 8.225E-01 | 8.148E-01 | 8.769E-01 | 7.775E-01 | 7.615E-01 | 8.548E-01 | 8.680E-01 |
| Training time (s) | 65 | 54 | 52 | 73 | 80 | 63 | 57 | 66 |

*3.3. Hyper-parameters tuning*

Tuning hyper-parameters in improving prediction accuracy plays a significant role in various domains [22, 23, 24] and enhances the performance of multi-modal deep learning models [25]. Indeed, some of the most important hyper-parameters, learning rate and batch size, are essential in deep CNNs because the learning rate determines the optimisation algorithm's step size during training to update the model's parameters. A higher learning rate can lead to more rapid convergence through training steps, but it may also yield the optimisation to surpass the optimal solution and evolve inconsistently [26]. Conversely, a lower learning rate can bring slower convergence, despite most cases leading to a more robust and accurate representation. Consequently, picking a reasonable learning rate is crucial for acquiring optimal performance in deep CNNs. Furthermore, the batch size is associated with the number of training examples that the model functions at once via each training iteration. Larger batch sizes can achieve training with lower iterations; however, it demands more memory and may overfit the model to the training data [27]. To prevent the overfitting issue, a smaller batch size might be useful and make a better generalisation ability. Hence, tuning the batch size is critical for balancing training efficiency and model accuracy.

As hyper-parameter learning rate and batch size are often interdependent and changing one can affect the other's optimal value, we tested various combinations of both in order to evaluate the performance of Inception-ResNet-V2.

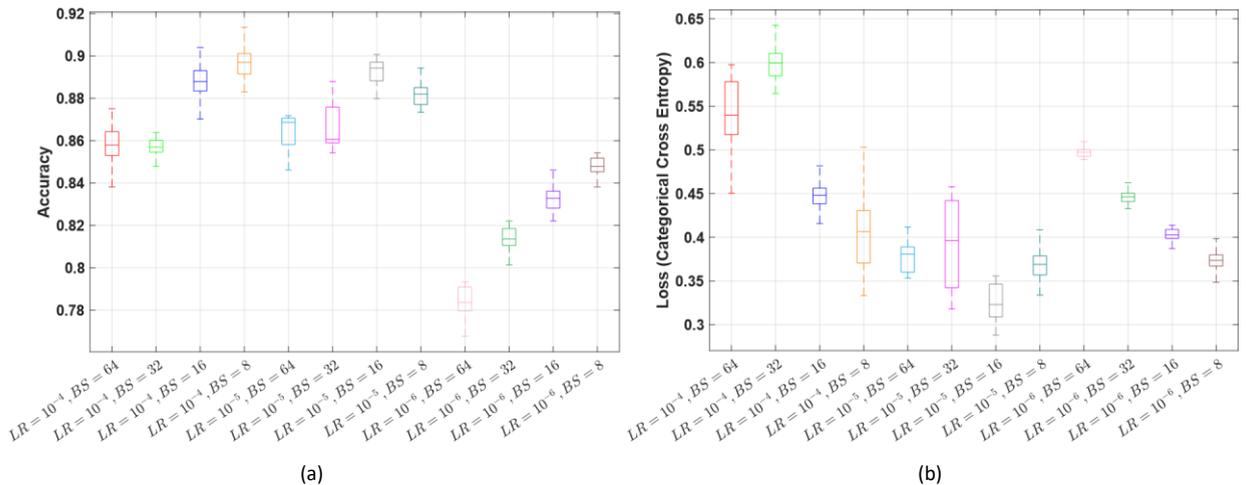

Fig. 5: Hyper-parameters impact on the performance of Inception-ResNet-V2 for classifying Pneumonia from normal cases (a) accuracy, and (b) loss validation analysis.

As can be seen in Figure 5, the highest accuracy was obtained with a high learning rate ($LR = 10^{-4}$) and small batch size ($BS = 8$). Indeed, the primary observation from this figure is that a larger batch size may require a higher learning rate to assure convergence; for example, a combination of $BS = 64$ and $LR = 10^{-4}$. Thus, tuning both hyper-parameters simultaneously can lead to better performance and faster convergence during training.

Tuning the number of trainable layers in deep CNNs is essential for preventing overfitting, improving learning, balancing computational requirements, and operating transfer learning effectively [19]. Training a deep CNN with a large number of trainable layers can impose a severe computational expense and be time-consuming. Consequently, restricting them can decrease the model's computational necessities and training time. Furthermore, many case studies have reported more inclined to overfitting for using Deep CNNs with many trainable layers. Thus, tuning the number of trainable layers can help prevent overfitting and improve the model's generalisation ability [28]. Figure 6 reflects the impact of various trainable layer numbers on the performance of Inception-ResNet-V2 in terms of (a) accuracy and (b) loss validation. We can see that the highest accuracy was achieved for $TL = 30$; however, the accuracy variance is higher than other trainable layers' numbers. To achieve a great balance between accuracy and loss validation, $TL = 20$ can be the best option as it provides the lowest average loss and highest accuracy. The Confusion Matrix (Figure 6 c) is a robust evaluation instrument for machine learning models tasked with multi-class classification problems. It also helps comprehend the nature of the model's errors, like misclassifications and offers a lucid visual representation of the classification outcomes. We can see the high ability of Inception-ResNet-V2 to classify Pneumonia samples from the normal case; however, the number of samples which classified wrongly is considerable and should be improved.

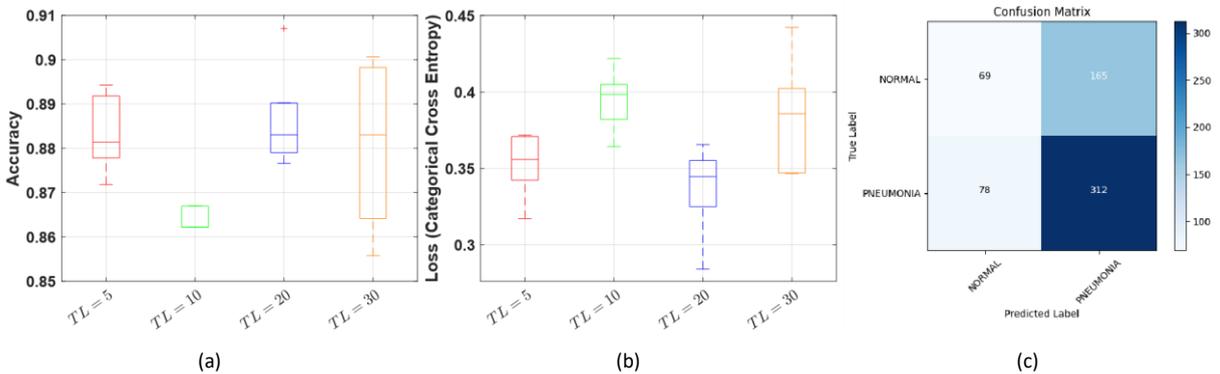

**Fig. 6:** The number of trainable layers (*TL*) of Inception-ResNet-V2 for fine-tuning and classifying Pneumonia from normal cases (a) accuracy, (b) loss validation analysis, and (c) confusion matrix.

4. **Conclusion and Future Work**

This article presents a comparative study of the Inception-ResNet deep learning model's performance in diagnosing pneumonia from chest radiographs. The study employs Mendeley's chest X-ray images dataset, which comprises 5856 2D images, including both Viral and Bacterial Pneumonia X-ray images. We compare the Inception-ResNet model with seven other state-of-the-art convolutional neural networks (CNNs), and the experimental results demonstrate the Inception-ResNet model's superiority in feature extraction and computational efficiency. Furthermore, we investigated the impact of transfer learning with fine-tuning in improving the performance of deep convolutional models. In terms of pneumonia classification accuracy, the findings show that Inception-ResNet-V2 surpassed ResNet152V2, MobileNet-V3 (Large and Small), EfficientNetV2 (Large and small), InceptionV3, and NASNet-Mobile at 2.6%, 6.5%,

7.1%, 13%, 16.1%, 3.9%, and 1.6%, respectively. Furthermore, Inception-ResNet excels in extracting essential features from Viral and Bacterial Pneumonia X-ray images due to combining the strengths of both Inception and ResNet architectures, leveraging the benefits of both. Thus, we can see a substantial improvement in validation loss between 9% and 45% compared with EfficientNetV2 and NASNet-Mobile. This study provides valuable insights into applying deep learning models for pneumonia diagnosis and highlights the potential of the Inception-ResNet model in this domain. In future work, it would be beneficial to extend the study to include out-of-distribution scenarios, where the test data is acquired from a different medical department than the training data. This would allow us to evaluate the model's performance and generalisation ability in more diverse and realistic settings.